\documentclass{article} 
\usepackage{iclr2016_conference,times}
\usepackage{hyperref}
\usepackage{url}
\usepackage{bm}
\usepackage{color}
\usepackage{graphicx}
\usepackage{amsmath}
\usepackage{amssymb}
\usepackage{caption}
\usepackage{subcaption}

\title{How far can we go without convolution: Improving fully-connected networks}

\author{Zhouhan Lin \& Roland Memisevic \\  
Universit\'e de Montr\'eal \\
Canada \\
\texttt{\{zhouhan.lin, roland.memisevic\}@umontreal.ca} \\
\And
Kishore Konda \\
Goethe University Frankfurt \\
Germany \\
\texttt{konda.kishorereddy@gmail.com} \\
}

\begin{document}
\maketitle

\begin{abstract}
We propose ways to improve the performance of fully connected networks. 
We found that two approaches in particular have a strong effect on performance: 
linear bottleneck layers 
and unsupervised pre-training using autoencoders without hidden unit biases. We show 
how both approaches can be related to improving gradient flow and reducing sparsity 
in the network. 
We show that a fully connected network can yield approximately $70\%$ classification 
accuracy on the permutation-invariant CIFAR-10 task, which is much higher 
than the current state-of-the-art.  
By adding deformations to the training data, the fully connected network achieves $78\%$ accuracy, which is just $10\%$ short of a decent convolutional network. 
\end{abstract}

\section{Introduction}
Convolutional neural networks (CNN) have had huge successes since \cite{krizhevsky2012imagenet}, especially in computer vision applications. 
The main computational idea behind these, \emph{weight sharing}, 
is unfortunately not biologically plausible, and it does not map nicely to 
simple, densely parallel hardware designs, which may one day yield lower-energy, 
and more efficient ways to run neural networks. 
The reason is that weight sharing requires long-range communication, 
for example, when distributing derivatives, during learning. 

In this paper, we explore the performance that one can possibly achieve with 
a neural network without convolutions, in other words, a neural network learning 
on ``permutation invariant'' classification tasks. 
We use these as a test-bed for studying alternative architectural design choices beside 
convolution, which are biologically plausible and potentially more hardware friendly.
Studying these design choices is relevant also because many tasks and sub-tasks 
do not come with any local translation invariance, making them not amenable to 
convolution (an example being the hidden layer in a recurrent neural network). 

Two architectural choices we found to have a strong impact on performance are 
(i) linear bottleneck layers 
and (ii) pre-training using autoencoders whose hidden units have 
no biases. 
Taken together, these two approaches allow us a network to achieve state-of-the-art 
performance on the permutation-invariant CIFAR-10 task \footnote{An example implementation that generates the state-of-the-art on this task is available at \url{https://github.com/hantek/zlinnet}}, and 
by adding deformations (thus removing the permutation invariance requirement) 
it achieves performance close to the range of that achieved by CNNs.

\subsection{Sparsity in neural networks}
Both approaches can be viewed from the perspective of sparsity in a neural network. 
Sparsity is commonly considered as a desirable property in machine learning models, 
and it is often encouraged explicitly as a regularizer. 
Unfortunately, in deep, multilayer neural networks sparsity comes at a price: 
in common activation functions, such as sigmoid or ReLU, zero (or almost zero) 
activations are attained at values where derivatives are zero, too. This prevents derivatives 
from flowing through inactive hidden units, and makes the optimization difficult. 
Stated in terms of the vanishing gradients problem (e.g., \cite{hochreiter2001gradient}) 
this means that for a ReLU activation many Jacobians are diagonal matrices containing 
many zeros along the diagonal. 

To alleviate this problem, recently several activation functions were proposed, 
where zero derivatives do not, or are less likely to, occur.  
In the PReLU activation function \citep{he2015delving}, for example, the zero-derivative 
regime is replaced by a learned, and typically small-slope, linear activation. 
Another approach is the Maxout activation \citep{goodfellow2013maxout}, which is defined as 
the maximum of several linear activations. 
Accordingly, preventing zero-derivatives this way was shown to improve the optimization 
and the performance of multilayer networks. 
Unfortunately, these methods solve the problem by giving up on sparsity altogether, 
which raises the question whether sparsity is simply not as desirable as widely assumed or whether the benefit of the optimization outweigh any detrimental effect on sparsity. 
Sparsity also plays a crucial role in unsupervised learning (which can also be 
viewed as a way to help the optimization \citep{saxe2013exact}), where these activations have 
accordingly never been successfully applied. 

This view of sparsity motivates us to revisit a simple, but as we shall show surprisingly 
effective, approach to retaining the benefits of sparsity without preventing gradient-flow. 
The idea is to inter-leave linear, low-dimensional layers with the sparse, 
high-dimensional layers containing ReLU or sigmoid activations. 
We show that this approach outperforms equivalent PReLU and Maxout networks on the fully supervised, permutation invariant CIFAR-10 task. 

A second detrimental side-effect of sparsity, discussed in more detail in \cite{konda2014zero}, is that for a ReLU or sigmoid unit to become sparse it typically learns strongly negative bias-terms. In other words, while sparse activations can be useful in terms of 
the learning objective (for example, by allowing the network to carve up space 
into small regions) it forces the network to utilize bias terms that are not optimal 
for the \emph{representation} encoded in hidden layer activations. 
\cite{konda2014zero}, for example, suggest bias-terms equal to zero 
to be optimal and propose a way to train an autoencoder with zero-bias hidden units.  
We suggest in Section~\ref{section:activepathorthogonolization} that 
pre-training a network with these autoencoders may be viewed as a way to 
orthogonalize subsets of weights in the network, 
and show that this yields an additional strong performance improvement.

We shall discuss the motivation for linear bottleneck layers in the next section and pre-training using autoencoders in Section~\ref{section:activepathorthogonolization}, followed by experimental results in Section~\ref{section:experiments}.

\section{Linear Bottleneck Layers}
\label{section:composing}
One drawback of sparsity in a deep network is that it amounts to data scarcity: 
a weight whose post-synaptic unit is off $80\%$ of the time will effectively get to see 
only $20\%$ of the training data. 
In lower layers of a convolutional network, this problem is compensated by weight-sharing,  
which increases the effective number of training examples (patches in that case) per weight by a 
factor of several thousand. In a fully connected layer (more precisely, a layer without 
weight sharing) such compensation is not possible and the only way to counter this effect 
would be by increasing the training set size. 

\begin{figure}[!h]
\begin{center}
    \begin{tabular}{cc}
    \includegraphics[width=0.5\textwidth]{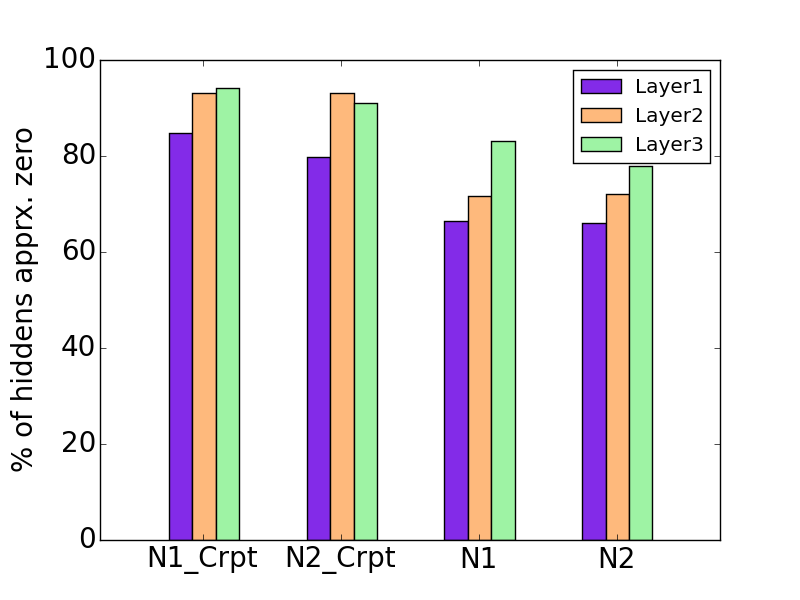}
    &\hspace{-10pt}\includegraphics[width=.5\textwidth]{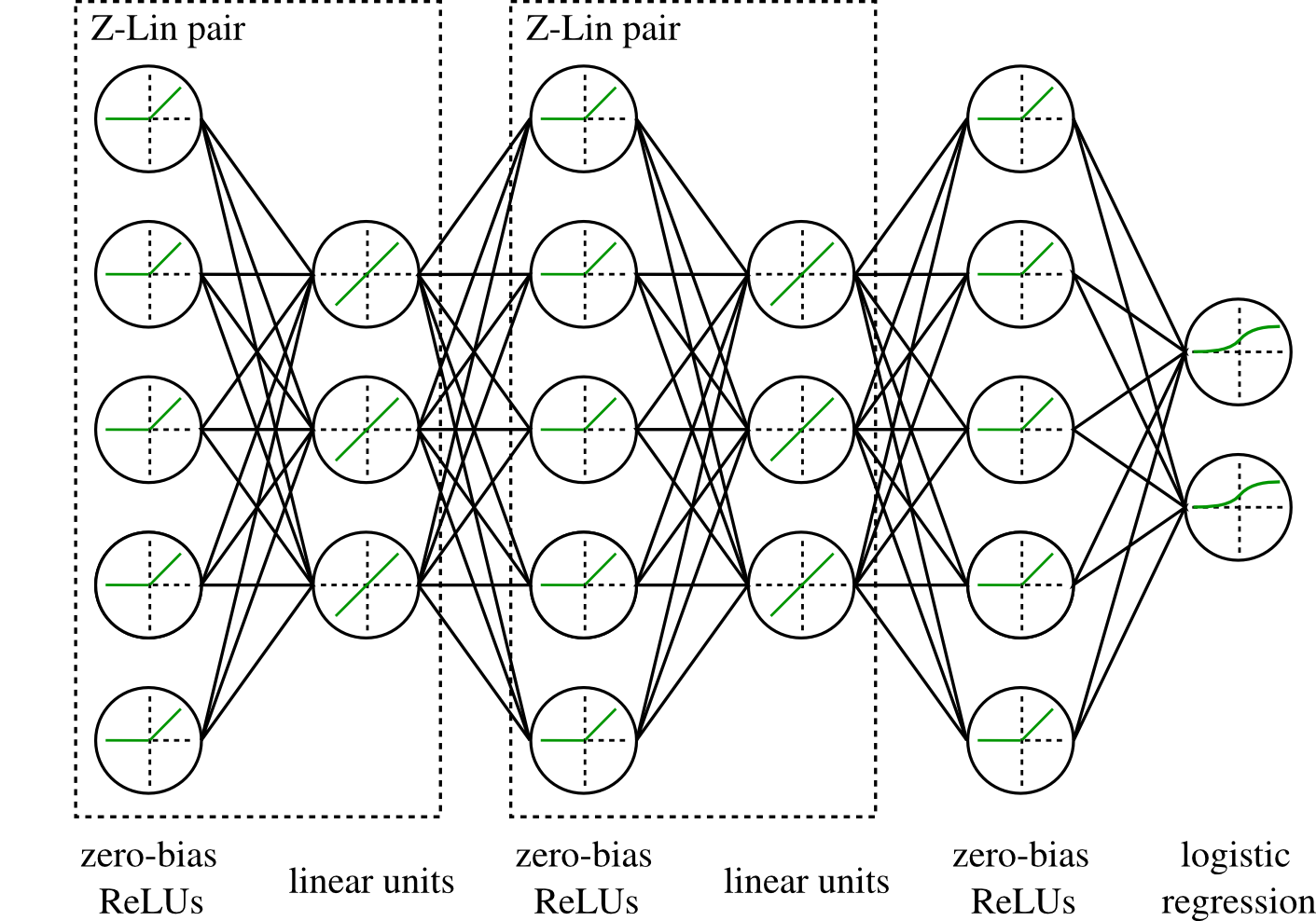}
    \end{tabular}
\caption{(left) The sparsity in different layers of two MLPs N1(1000-2000-3000 units) and N2(2000-2000-2000 units) trained with and without dropout on CIFAR-10 dataset.
N1\_Crpt, N2\_Crpt: Experiments with dropout. 
(right) Illustration of a network with linear bottleneck layers.}
\label{zlin}
\end{center}
\end{figure}

Sparsity is a common and stable effect in neural networks containing ReLU or sigmoid units,
and it can be related to the fact the biases are driven to zero in response to regularization 
(eg., \cite{konda2014zero}).
Figure \ref{zlin} (left) shows the sparsity level attained after training ReLU MLPs. \footnote{Two MLPs (1000-2000-3000 units and 2000-2000-2000 units, respectively), trained for $501$ epochs on CIFAR-10. Each network was trained once with dropout and once without. For the experiments with dropout, an input noise level of $0.2$ and hidden noise level of $0.5$ was used. }
The plots show that the sparsity of the hidden presentation increases 
with the depth of the network and increase further in the presence of dropout.

\subsection{Linear Layers as Distribution Reshaping}   \label{distremap}
It is well-known that different activation functions work best when the input is 
in a reasonable scale, which is the motivation for using pre-processing by mean subtraction and 
scaling, for example. 
Furthermore, incorporating them into hidden unit activation, can yield further performance 
improvements \citep{ioffe2015batch}.

Here, we focus on the distribution over the outputs of a unit. 
The output of a linear neuron $Y$ with inputs $\vec{X}=(X_{1}, X_{2}, ..., X_{i}, X_{N})$ 
is given by
$
Y = \sum_{i}^{N}{{w}_{i}{X}_{i}} + b,
$
where ${w}_{i}$ is the entry in the weight vector corresponding to input node $i$, $b$ is the bias of the linear neuron, and $N$ denotes the input dimension. 
Assume the data fed into each input node ${X}_{i}$ is independent 
and has a finite mean ${\mu}_{i}$ and variance ${\sigma}_{i}^{2}$. 
In the net input to a neuron, 
the mean and variance of each ${w}_{i}{X}_{i}$ term 
are tuned by its corresponding weight values:
\begin{equation}
{\hat{\mu}}_{i} = {w}_{i}{\mu}_{i}; \quad \quad
{\hat{\sigma_{i}}}^{2} = {w}_{i}^{2}{\sigma}_{i}^{2}
\end{equation}
According to the Lyapunov theorem \citep{degroot1986probability}, if a sequence of random variables with 
finite mean and variance are independent but not necessarily identically distributed, 
the distribution of the sum:  
\begin{equation}
S = \Big( \sum_{i=1}^{N}{{w}_{i}{X}_{i}} - \sum_{i=1}^{N}{ {\hat{\mu}}_{i} } \Big){\Big({\sum_{i=1}^{N}{ {\hat{\sigma_{i}}}^{2} }}\Big)}^{-\frac{1}{2}}
\end{equation}
tends to standard Gaussian distribution for $N\xrightarrow{}\infty$. 
If we write $Y$ in terms of $S$, that is, 
$
Y = {\left({\sum_{i=1}^{N}{ {\hat{\sigma_{i}}}^{2} }}\right)}^{\frac{1}{2}}S + b + \sum_{i=1}^{N}{ {\hat{\mu}}_{i} }
$
we see that $Y$ approximates a Gaussian distribution whose mean is $b + \sum_{i=1}^{N}{ {\hat{\mu}}_{i} }$ and whose variance equals to $\sum_{i=1}^{N}{ {\hat{\sigma_{i}}}^{2} }$, when the input dimension $N\xrightarrow{}\infty$. 
For mean-centered data, we have ${\hat{\mu}}_{i}\approx0$. Thus the actual mean value of $Y$ is merely dominated by $b$ in which case the p.d.f. of $Y$ is:  
\begin{equation}
p_{ Lin }\left( y \right) \approx \frac { 1 }{ \sqrt { 2\pi \sum _{ i=1 }^{ N }{ { \hat { \sigma _{ i } }  }^{ 2 } }  }  } { e }^{ -\frac { { \left( y-b \right)  }^{ 2 } }{ 2 \sum _{ i=1 }^{ N }{ { { \hat { \sigma  }  }_{ i } }^{ 2 } }  }  }
\end{equation}
This form of asymptotic distribution holds regardless of the weight ${w}_{i}$. That means, no matter how the network is trained, or even not trained, the asymptotic distribution of output of a linear unit tends to be Gaussian.

\begin{figure}[!h]
\begin{center}
\centerline{\includegraphics[width=\textwidth]{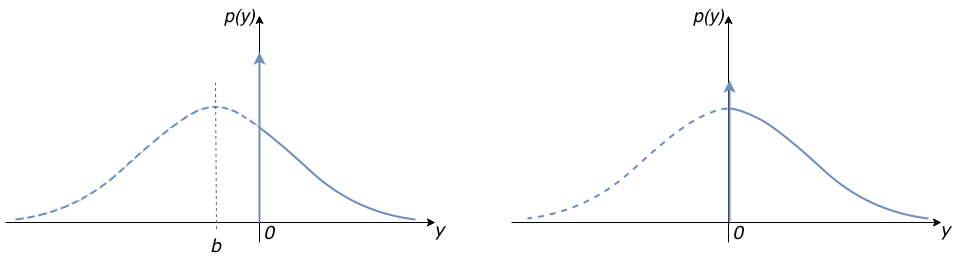}}
\caption{P.d.f. of ReLU output (left) and zero-bias ReLU output (right). The arrow at zero indicates a delta spike, and the dashed part stands for the probability mass absorbed into that delta spike. 
}
\label{pdf}
\end{center}
\end{figure}

Since the pre-hidden activation of ReLU is linear it tends to be Gaussian as well. The ReLU activation then simply sets all negative values to zero which yields: 
\begin{equation}  \label{relupdf}
p_{ ReLU }\left( y \right) = \begin{cases}  \int _{ -\infty  }^{ 0 }{ p_{ Lin }\left( y \right) \mathrm{d}y} \cdot \delta(0), y \le 0 \\ p_{ Lin }\left( y \right), y > 0 \end{cases}
\end{equation}
where $\delta(0)$ is the Dirac delta. The distribution has a delta spike at zero and a Gaussian tail at its positive end (Figure~\ref{pdf}). 

\begin{figure}[!h]
\begin{center}
    \begin{subfigure}[b]{0.3\textwidth}
    \includegraphics[width=\textwidth]{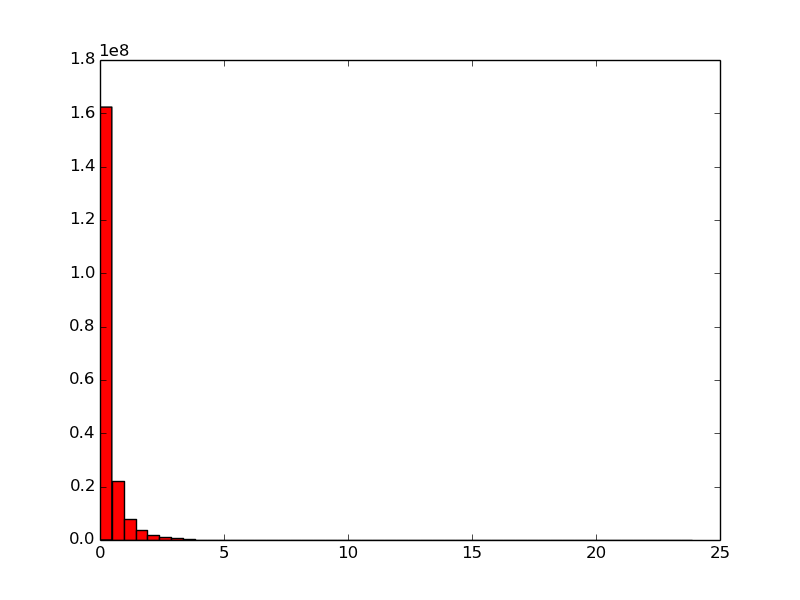}
    \caption{zero bias ReLU activation}
    \label{zaedistro}
    \end{subfigure}
    \begin{subfigure}[b]{0.3\textwidth}
    \includegraphics[width=\textwidth]{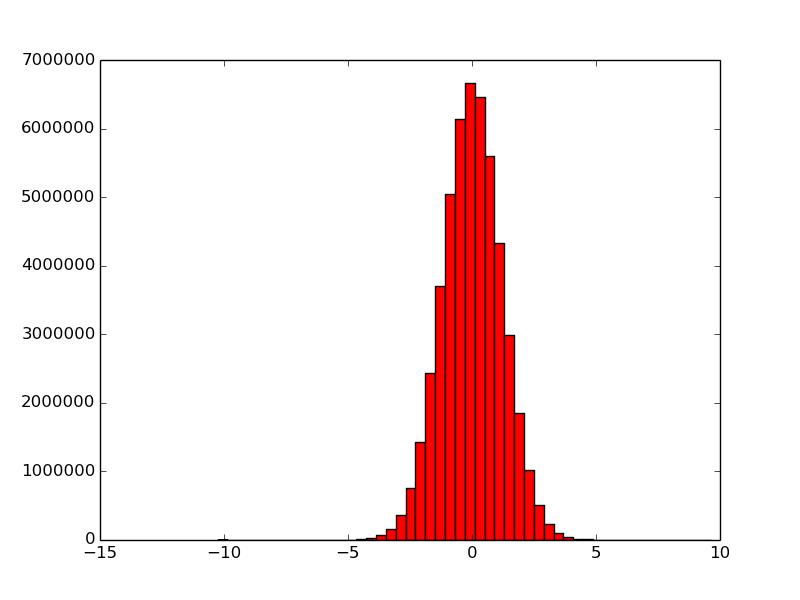}
    \caption{linear layer activation}
    \label{lindistro}
    \end{subfigure}
    \begin{subfigure}[b]{0.3\textwidth}
    \includegraphics[width=\textwidth]{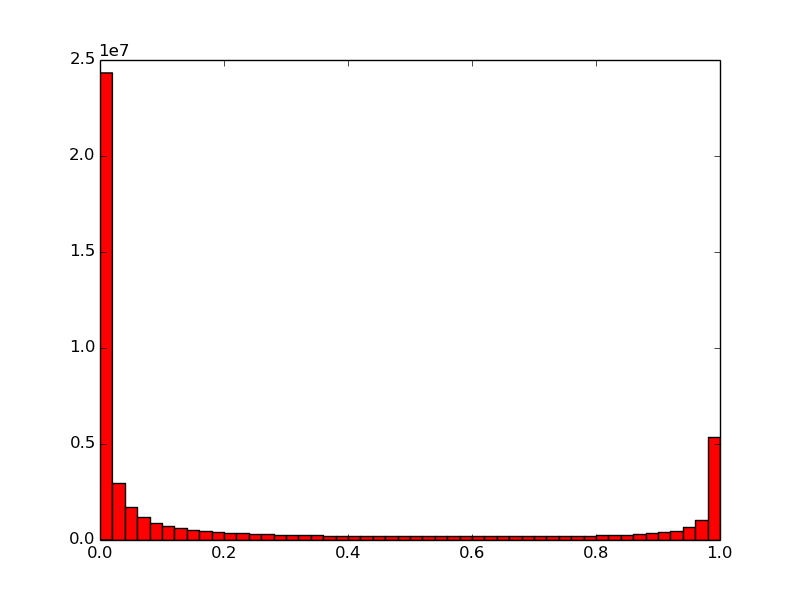}
    \caption{Sigmoid activation}
    \label{sigdistro}
    \end{subfigure}
    \caption{Histogram of output activation with different types of activation function.}
\label{activation}
\end{center}
\end{figure}

Since the bias controls the intensity of the $\delta(0)$ spike, it controls the sparsity of output representation (left plot). 
The observation that biases tend to zero motivated \cite{konda2014zero} to introduce 
a ``zero-bias'' activation function  which uses a fixed threshold followed by 
linear activation.  
Typically while using zero-bias ReLU, a threshold of 1 is introduced during pre-training stage, and set back to zero while training its subsequent layers and fine-tuning. 
The distribution of pre-hidden activity of a zero-bias ReLU stretches equally on both sides of zero. 
As a result, for zero-bias ReLU activation half of the probability 
mass concentrates on a delta spike located at zero, as illustrated it 
Figure \ref{pdf} (right).
While batch normalization alone will push the mean and variance into the optimal range for a 
subsequent ReLU unit, it will not resolve the issue that the distribution is peaked at the 
negative bias. In fact, the estimate of mean and variance will suffer from the presence 
of a highly non-Gaussian distribution.
Typical histograms over hidden unit activations for several activation functions are shown in Figure~\ref{zaedistro}.

\subsection{Derivatives in the Presence of Linear Layers}
We now consider the derivatives of a ReLU network. 
The activation of the $i$-th layer is given by: 
\begin{equation}
\vec{H_{i+1}} = R(w_i \vec{H_i} + \vec{b_i} ), 
\end{equation}
where $R()$ is the element-wise activation function. 
The back-propagated updates on $w_i$ will be:
\begin{equation}
\Delta w_i = \delta \circ R^{'}\left( \vec{H_{i+1}} \right) \cdot \vec{H_i}
\end{equation}
where $\delta$ stands for the down flowing error signal coming from the upper layer, $\circ$ 
stands for element-wise multiply, and $\cdot$ is the product of a vertical vector and 
a horizontal vector. 
As discussed in Section \ref{distremap}, at least 50\% (usually much more in practice due to 
negative biases) 
of the values in the representation $H_i$ and $H_{i+1}$ are typically 
equal to zero (cf., Figure~\ref{zlin}, left) 
making a large fraction of the entries of $\Delta w_i$ per training case zero. 
If we introduce a linear layer between two ReLU layers, 
\begin{equation}  \label{wlwi}
\vec{H_{l}} = w_l \vec{H_i} + \vec{b_l}\quad
\vec{H_{i+1}} = R\left( w_{i} \vec{H_{l}} + \vec{b_{i}} \right)
\end{equation}
where $\vec{H_{l}}$ stands for the output of linear layer, and $w_{l}$, $b_{l}$ are the weights and biases in the linear layer, we obtain updates of the form: 
\begin{equation}
\Delta w_{l} = \delta_{l} \cdot \vec{H_i}\;,\; \; \; \;
\Delta w_i = \delta \circ R^{'}\left( \vec{H_{i+1}} \right) \cdot \vec{H_{l}}
\end{equation}
where $\delta_{l}$ is the error signal passed to the linear layer, $\delta_{l} = \delta w_i$. Since the linear layer representation $\vec{H_{l}}$ is dense, both $\Delta w_{l}$ and $\Delta w_i$ become denser: only half of the values in these two update matrices are zeros.
More importantly, there always exists a simple ReLU layer that is equivalent to a ReLU/linear combination 
because any linear layer can be absorbed into the weights of the ReLU layer: 
\begin{equation}
w = w_i w_{l}, \quad b = w_i b_{l} + b_i
\end{equation}
Then, the equivalent update on $w$ becomes:
\begin{equation} \label{intrinsic}
\Delta w = \Delta w_i \Delta w_{l} + w_i \Delta w_{l} + \Delta w_i w_{l}
\end{equation}

Even if half of the values in $\Delta w_i$ and $\Delta w_{l}$ are zero, their product is a dense matrix. The second and third term in Equation \ref{intrinsic} are also dense, 
so with a linear bottleneck layer, we actually obtain a dense update in an equivalent ReLU layer. 

\subsection{Reducing Parameters}
Besides helping with back propagation, it is important to note that linear layers also reduce the total number of parameters. 
Suppose a linear layer with $L$ units is inserted between two nonlinear layers with $N$ units each. The total number of parameters would become $2NL + L + N$. This is much less than the number of parameters that would result from directly connecting the two nonlinear layer, which would amount to $N^{2} + N$ parameters.
Convolutional network also reduces the number of parameters by convolution kernels. Note that for any trained convolutional network, we can always find a fully connected network with the same accuracy by expanding the convolutional kernels. 

\section{Pre-training and zero-bias activations}
\label{section:activepathorthogonolization}
It was suggested by \cite{saxe2013exact} that the benefit of unsupervised pre-training 
of a network using RBMs or autoencoders may result in weight matrices that are 
closer to orthogonal and thus less affected by vanishing gradients problems. 
It is interesting to note, however, that the sparsity-inducing negative biases 
yield reconstructions that are affine not linear and accordingly may not orthogonalize
weights after all \citep{konda2014zero}. 
This may be one of the reasons why the practical success of these pre-training schemes 
has been quite limited by comparison to fully supervised learning using back propagation. 

It is also important to note that due to sparsity, the number of active units 
in a layer is often smaller than that the same number in the layer below, so the 
``effective'' weight matrix for a given input example is not a square matrix.  
Rather than simply orthogonal weight matrices, we should be looking for networks where 
hidden units which tend to be active on the same inputs have weights that tend to 
be orthogonal to one another. 
In other words, we should be looking for ``orthogonal active paths'' through the
network rather than overall orthogonal weight matrices. 

In order to obtain hidden units with linear not affine encodings 
\cite{konda2014zero} introduce ``zero-bias autoencoders'' (ZAE) whose activations 
are threshold-ed when pre-trained by minimizing the autoencoder reconstruction error, 
and whose thresholds are removed when the weights are used for classification or 
for initializing a feed forward network. 
As discussed in \cite{konda2014zero}
minimizing squared error under the linear encoding should encourage weights 
corresponding to hidden units that tend to be ``on'' together to orthogonalize (because 
the orthogonal projection is what minimizes reconstruction error). 
\cite{konda2014zero} reported decent classification performance in various supervised 
tasks, but found only a weak improvement over standard autoencoders when 
used to initialize a single-hidden-layer MLP. 

The view of a zero-bias activation function as a way to orthogonalize weights suggests
especially deep networks to profit from ZAE pre-training, and so we investigated the
performance of ZAE-pretrained networks with many hidden layers, as well as in 
conjunction with interleaved bottleneck layers (as discussed 
in Section~\ref{section:composing}). We found them to yield a 
separate, and significant, performance improvement in fully connected networks.

\section{Experiments}
\label{section:experiments}

\subsection{CIFAR-10}  \label{pvz}
The CIFAR-10 dataset is a subset of the 80 Million Tiny Images \cite{torralba200880}, and contains $10$ balanced classes. It provides a training set with $50000$ samples and a test set of $10000$ samples. Each sample is a color image with $32\times32$ RGB pixels.
We first compare mixed models with linear bottleneck layers and ordinary networks. We compare two different activation functions: ReLU and zero-bias ReLU \citep{konda2014zero}. 
Comparison are based on classification accuracy. All the experiments in this subsection share the same pre-processing pipeline and the same type of classifier. For pre-processing, the raw data is contrast normalized and centered to have zero mean, followed by PCA whitening, retaining 99\% of the variance. 

\begin{figure}[!h]
\begin{center}
    \begin{subfigure}[b]{0.49\textwidth}
    \includegraphics[width=\textwidth]{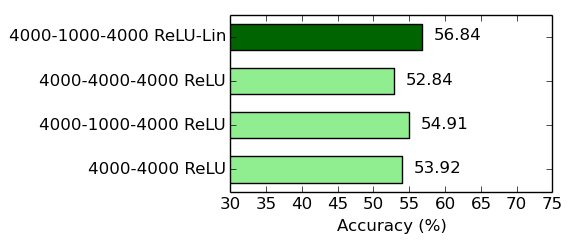}
    \caption{ReLU-Lin network and various networks with pure ReLU activation.}
    \label{pvmrelu}
    \end{subfigure}
    \begin{subfigure}[b]{0.49\textwidth}
    \includegraphics[width=\textwidth]{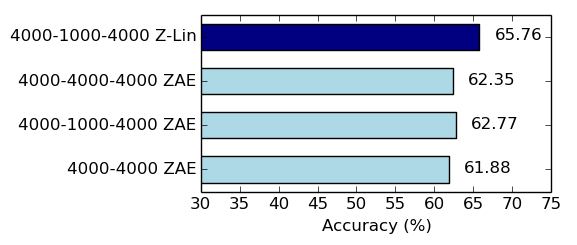}
    \caption{Z-Lin network and various networks with pure Zero-bias ReLU activation.}
    \label{pvmz}
    \end{subfigure}
    
    \begin{subfigure}[b]{0.49\textwidth}
    \includegraphics[width=\textwidth]{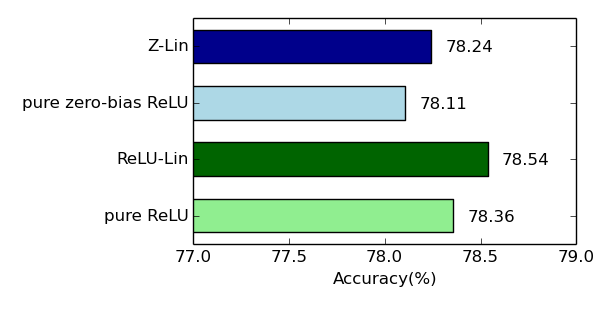}
    \caption{Z-Lin, zero-bias ReLU, ReLU-Lin and ReLU network trained on HIGGS dataset.}
    \label{fig_higgs}
    \end{subfigure}
\caption{Comparing bottlenecked network with its various counterparts. Models compared in (a) and (b) are trained on CIFAR-10, while those in (c) are trained on HIGGS dataset.}
\label{purevsmix}
\end{center}
\end{figure}

We trained 3 ReLU networks with different configurations, and compare their results 
with the ReLU-Lin network $4000{\rm ReLU}-1000{\rm Linear}-4000{\rm ReLU}$, that interleaves a linear 
layer between two ReLU layers. 
All networks in Figure \ref{pvmrelu} are trained by supervised back propagation using stochastic 
gradient descent with 0.9 momentum. 
The ReLU-Lin network outperforms all the pure ReLU networks and reaches an accuracy 
of 56.84\%. 
We also found that the ReLU-Lin network tends to be more stable than the pure ReLU networks: 
if we pre-train these networks layer-wise, then for the ReLU networks the majority of the 
units in higher layers become ``dead'', whereas the ReLU-Lin network is still stable 
for unsupervised pre-training without dead units at the second ReLU layer. 

We repeat the same experiment by substituting ReLU with ZAE (Figure \ref{pvmz}). 
The Z-Lin network is configured as $4000{\rm Z}-1000{\rm Linear}-4000{\rm Z}$ (same as in Figure \ref{zlin}, 
but with only one Z-Lin pair). 
Since ZAE does not suffer from the problem of dead units that much as ReLU, 
all the networks are first layer-wise pre-trained in an autoencoder, and then fine-tuned 
with stochastic gradient descent. During pre-training the linear layer, we have 1.0 weight 
decay added to the cost. Similar to the case of ReLU, it is also observed that introducing 
linear bottleneck layers makes the stacked deeper model outperform its shallow counterpart. 
Having a linear inserted in the middle makes the Z-Lin model outperform all the other models, 
yielding an accuracy of 65.76\%.

We also compare the bottleneck layer structure with other related modifications like PReLU \citep{he2015delving} and Maxout \citep{goodfellow2013maxout}, the bottleneck structure performs better than both of them, if we apply it on the same network architecture. In comparison to batch normalization \citep{ioffe2015batch} (but on fully connected networks) on 
the other hand, their improvements are about the same.

\subsection{The HIGGS Dataset}  \label{higgs}
The HIGGS dataset has $11$ million samples with $28$ dimensions. 
The first $21$ features are kinematic properties measured by particle detectors in 
a particle accelerator, and the remaining $7$ features are functions of the first 21 features. 
Thus the dataset itself is permutation invariant.
The task is to decide if or not a given sample corresponds to a Higgs Boson. 

We tried both ReLU and ZAE with/without linear bottleneck layers on this dataset. 
Similar to before, for each model PCA whitening retaining $99\%$ of the variance 
is used for pre-processing. 
(corresponding to $27$ principle components). 
We use the same model size for four different models that we compare (zero-bias ReLU, Z-Lin, ReLU, and ReLU-Lin). 
The structure is $27-800-100-800-100-2$ for all, so the models differ by using different 
activation functions. 
We train all the models using SGD with momentum, and tune learning rates individually. 
We do not use any pre-training. The results are shown in Figure \ref{fig_higgs}, which also confirm the effectiveness of the bottleneck layers, albeit not as pronounced as on the CIFAR-10 data. 
Also, zero-bias units do not yield an improvement here. 

\subsection{Reducing Parameters}
In this subsection we explore how the network's performance is affected by 
increasingly reducing the number of parameters, using bottlenecks of different sizes.

\begin{figure}[!h]
\begin{center}
\centerline{\includegraphics[width=.5\textwidth]{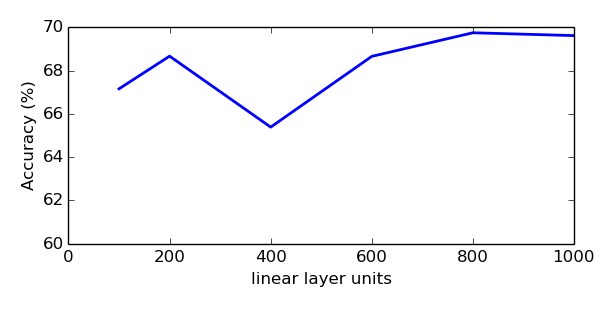}}
\caption{Classification accuracy w.r.t different linear layer size. }
\label{reduceparam}
\end{center}
\end{figure}

We define a network by stacking $3$ Z-LIN pairs, plus a ZAE layer and a logistic regression classifier. Each ZAE layer has $4000$ hidden units. We reduce the linear layer size from $1000$ down to $100$ units. Training involves dropout, pre-training and fine-tuning. The results are shown in Figure~\ref{reduceparam}. We observe that even with a hidden layer of size $100$, which has only $1/20$ of the parameters, the network still works reasonably well and does not loose too much accuracy.

A 7-layer fully connected network with $4000$ units each layer has around $112$ million parameters. With linear layers, the smallest model in Figure \ref{reduceparam} has only around $2.44$ million parameters. By comparison, a typical convolutional network yielding an accuracy higher than 80\% on CIFAR10 would have around $3.5$ million parameters.

\subsection{Achieving state-of-the-art on Permutation Invariant CIFAR-10}
We compare the performance of the Z-Lin model with other published methods on the CIFAR-10.
For our method, pre-processing steps are the same to Section \ref{pvz}. 
For the Z-Lin network, each ZAE layer $4000$ hidden units, and each linear autoencoder 
has $1000$ hidden units. We use logistic regression on top of the last ZAE layer for classification. 
Following \cite{konda2014zero}, the threshold of all ZAEs are fixed at $1.0$ during pre-training, 
and set to $0$ while training the subsequent layers and performing fine-tuning. 
As before, we subtract the mean value and normalize the activations to have a standard 
deviation of $1.0$ in all layers. 

We train the networks using stochastic gradient descent with $0.9$ momentum and 
decreasing learning rate. The learning rate is set to $0.001$ for the ZAEs and to $0.0001$ 
for the linear autoencoder. 
Weight decay is used for the linear autoencoder and for the logistic regression layer.  
The latter is trained using nonlinear conjugate gradients. 
After pre-training, we use a tiny learning rate of $5\times10^{-6}$ to fine-tune the 
whole network, which yields an overall accuracy of $65.7\%$. This already outperforms 
all previous published permutation invariant CIFAR-10 results, the next best-performing of 
which are $63.1\%$ \citep{fastfood}, and $63.9\%$ \citep{konda2014zero}.

By adding dropout \citep{srivastava2014dropout} during pre-training and fine-tuning, and using 
a very deep Z-LIN network ($3$ Z-LIN pairs, plus a ZAE layer and logistic regression classifier, i.e., $4000{\rm Z}-1000{\rm Lin}-4000{\rm Z}-1000{\rm Lin}-4000{\rm Z}-1000{\rm Lin}-4000{\rm Z}-10$) the performance improves to $69.62\%$, 
which exceeds the current state-of-the-art on permutation invariant CIFAR10 by a very large margin. 

If we give up on permutation-invariance by using data augmentation 
(eg., \cite{krizhevsky2012imagenet}) but retain the use of a fully connected network, 
the performance improves much further. 
Here, we add flipping, rotation, and shifting to the original data during training of a 
$4000{\rm Z}-1000{\rm Lin}-4000{\rm Z}-10$ Z-LIN network pushing the performance to $78.62\%$. 
This is a much higher accuracy on CIFAR-10 achieved by any fully connected network we are 
aware of, and it is only about $10\%$ behind a well-trained convolutional network.

\begin{figure}[!h]
\begin{center}
\centerline{\includegraphics[width=.7\textwidth]{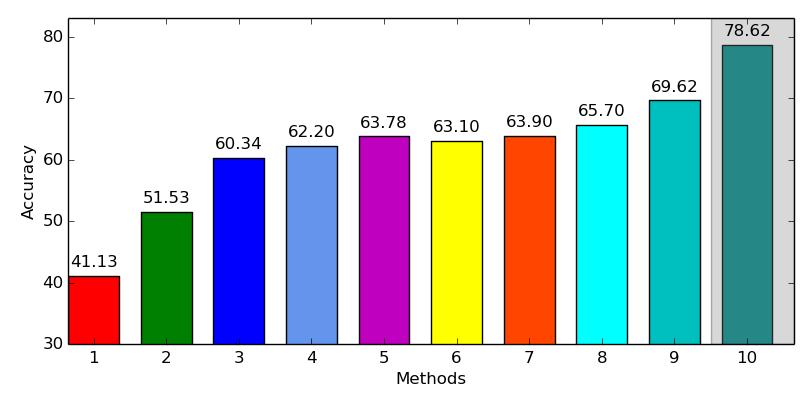}}
\caption{Test set accuracy of various methods. They are (from left to right):
1) Logistic Regression on whitened data;
2) Pure backprop on a 782-10000-10 network;
3) Pure backprop on a 782-10000-10000-10 network;
4) RBM with 2 hidden layers of 10000 hidden units each, plus a logistic regression;
5) RBM with 10000 hiddens plus logistic regression;
6) "Fastfood FFT" model \citep{fastfood};
7) Zerobias autoencoder of 4000 hidden units with logistic regression \citep{konda2014zero};
8) 782-4000-1000-4000-10 Z-Lin network trained without dropout;
9) 782-4000-1000-4000-1000-4000-1000-4000-10 Z-Lin network, trained with dropout
10) Z-Lin network the same as (8) but trained with dropout and data augmentation;
Results (1)-(5) are from \cite{krizhevsky2009learning}. The final one is distinguished with a grey background because it uses data augmentation.
}
\label{methods_compare}
\end{center}
\end{figure}

\section{Related Work}
The idea of a linear bottleneck layer is very old and has been used as early as 
the $1980$'s in the context of autoencoders (eg., \cite{baldi1989neural}). 
More recently, \cite{ba2014deep} used a linear bottleneck layer to factorize a single-layer 
network and showed that it helped speed up learning. An application of a linear bottleneck 
layer in the last layer of a neural network for dealing with high-dimensional outputs is described 
in \cite{sainath2013low} and \cite{xue2013restructuring}. 
In contrast to our work, in none of these methods is the goal to 
alleviate vanishing gradients and deal with sparsity, 
and (accordingly) they use just a single bottleneck 
layer in the network.  

Reshape distributions over activations using linear layers is also related to the recently introduced 
batch-normalization
trick \citep{ioffe2015batch}, in that it is also a way to adjust the distribution of inputs
to a subsequent layer. In contrast to that work, linear bottleneck layers not only adjust 
the mean and variance of the inputs to the subsequent layer, but reshape the whole \emph{distribution}.

\section{Discussion}
It is well known that a single-hidden-layer neural network can model any non-linear function under 
mild conditions \citep{funahashi1989approximate,cybenko1989approximation}. The intuition behind this observation 
is that the hidden layer carves up space into half-spaces, or ``tiles'', and the subsequent linear 
layer composes the non-linear function by combining different linear regions to produce the output. 
It is interesting to note that this view may suggest using a \emph{pair} 
of layers (a non-linear followed by a linear layer) to define the non-linear function, 
leading thus to interleaved linear/non-linear layers. 

The practical usefulness of this result is limited, however, because to approximate any given 
function it would require an exponentially large number of hidden units.  
In practice, this is one motivation for using multilayer networks
which compute a sequence of consequently more restricted, but tractable non-linear functions. 
Arguably, in the presence of enough training data and computational resources, 
wide hidden layers would still be preferable to narrow ones. However, in practice,  
wider hidden layers also entail more sparsity, which prevents the flow of 
derivatives. 
Also, as sparse activations propagate upwards through the network, 
they tend to proliferate, aggravating the problem in higher layers.

\subsubsection*{Acknowledgments}
The authors would like to thank the developers of Theano \citep{Bastien-Theano-2012}.
We acknowledge the support of the following agencies for research funding and
computing support: Samsung, NSERC, Calcul Qu\'{e}bec, Compute Canada,
the Canada Research Chairs and CIFAR.

\bibliography{zlin}
\bibliographystyle{iclr2016_conference}

\end{document}